# COVID-19 Pandemic Cyclic Lockdown Optimization Using Reinforcement Learning

Mauricio Arango[*], Lyudmil Pelov

*Oracle Corporation*

**ABSTRACT**

This work examines the use of reinforcement learning (RL) to optimize cyclic lockdowns, which is one of the methods available for control of the COVID-19 pandemic. The problem is structured as an optimal control system for tracking a reference value, corresponding to the maximum usage level of a critical resource, such as ICU beds. However, instead of using conventional optimal control methods, RL is used to find optimal control policies. A framework was developed to calculate optimal cyclic lockdown timings using an RL-based on-off controller. The RL-based controller is implemented as an RL agent that interacts with an epidemic simulator, implemented as an extended SEIR epidemic model. The RL agent learns a policy function that produces an optimal sequence of open/lockdown decisions such that goals specified in the RL reward function are optimized. Two concurrent goals were used: the first one is a public health goal that minimizes overshoots of ICU bed usage above an ICU bed threshold, and the second one is a socio-economic goal that minimizes the time spent under lockdowns. It is assumed that cyclic lockdowns are considered as a temporary alternative to extended lockdowns when a region faces imminent danger of overpassing resource capacity limits and when imposing an extended lockdown would cause severe social and economic consequences due to lack of necessary economic resources to support its affected population during an extended lockdown.

## 1. Introduction

This article examines the use of RL [1] to optimize cyclic lockdowns, which is one of the methods available for control of the COVID-19 pandemic. As originally described in [2] and [3], cyclic lockdowns are a control method in the epidemic control toolbox. During the first phase of the pandemic, stringent combinations of Non-Pharmaceutical Interventions (NPIs) were applied by governments to slow down the exponential spread of the disease and reduce epidemic indicators to low values. We assume a lockdown is the combination of multiple NPIs, including stay-at-home orders, school closures, and business closures, and that these NPIs are applied and released together. Across the world lockdown measures have generally been very successful and have helped save millions of lives as discussed in [4] and [5]. However, lockdowns have high economic and social costs, so in some regions it may be unsustainable to leave them in place for very long intervals, pressing governments to reduce restrictions. Yet the risk of new epidemic waves becomes high if the degree of unprotected social interaction increases too much during opening phases, which would require new deployment of lockdowns.

We assume cyclic lockdowns are used only as an alternative to extended lockdowns when a region's government, due to economic limitations, cannot provide sufficient support to its affected population. In cases where there is an urgent need to lower the rate of spread of the virus in order to avoid overpassing the available capacity of critical resources, such as ICU beds and ICU medical teams, and at the same time the economy cannot be completely closed, cyclic lockdowns can be a viable alternative as described in [6]. Cyclic lockdowns can be used to buy time to strengthen other critical processes needed to control the epidemic, including testing, contact tracing, isolation facilities, hiring of necessary medical teams, expanding medical equipment stockpiles, and deploying unified and effective community education programs on required social practices, including use of masks and distancing.

---

[*] Corresponding author email: mauricio.arango@oracle.com

The main contribution of this effort is the development of a tool to estimate the optimal timing for open and closed segments during cyclic lockdowns. We approach this problem by framing it as an optimal control problem and solving it using RL methods. From the RL point of view it is a sequential decision problem with a combination of public health and economic impact goals that need to be optimized. The tool comprises a simulation and optimization framework that integrates a dynamic epidemic model with an RL control agent. The tool helps answer questions such as: when would the first lockdown occur, what is the percentage of time spent in open and in lockdown mode, what is the average length in days for both the open and closed segments, and what is the maximum *effective reproduction number* [7], $R(t)$, that would avoid lockdowns.

## 1.1. Related Work

There is previous work in applying RL in epidemic control as described in [8] and [9]. However, to our knowledge there is no previous work in applying RL to optimize cyclic lockdowns. Also related, is work on using feedback control to guide the application of intervention measures [10]. Yet our work differs in that it provides *optimal* control and does so using RL methods. Cyclic lockdown strategies have been studied in [6] where predictable fixed open and closed cycle segment lengths are proposed (e.g. 4 days open and 10 closed). However, our work differs in that it optimizes cycles according to public health and economic goals.

Cyclic lockdown patterns studied in [2] and [3] use a basic heuristic feedback control method where a simple fixed rule is used to decide when to apply or release a lockdown. The fixed rule applies a lockdown when the ICU beds in use exceeds a high ICU bed threshold and releases the lockdown when the ICU beds in use goes below a low ICU bed threshold. This approach has two disadvantages: on the one hand, it can cause very large ICU bed usage spikes that sometimes exceed the total ICU bed capacity and on the other hand it produces long lockdown cycles. We refer to this basic method as the *baseline on/off control* method and contrast our solution with it. Our implementation of the baseline on/off control method used the same value for both the high and low ICU bed threshold.

## 1.2. Problem Framing

This work is structured as an *optimal control tracking* problem where the optimization goal is to bring one of the epidemic system variables, the number of ICU beds in use, as close as possible to a reference input which is the ICU bed threshold. The input control action is the epidemic's effective reproduction number [7], $R(t)$, which can only take two values, for example 1.5 (open) or 0.7 (lockdown). This type of on/off control is also referred to as *bang-bang* control. Instead of using conventional optimal control methods to find an optimal control policy function for $R(t)$, we use RL as described in [11], [12], [13], [14]. In contrast to conventional optimal control, RL does not require a model of the target system dynamics. With RL, a control policy function can be learned by an RL control agent interacting directly with the target system or with a simulator. In this case, although a dynamic model of the epidemic is available, it is used as a simulator and is decoupled from the RL control agent. This provides significant flexibility to modify either the simulation model or the RL control agent with minimal or no impact on each other.

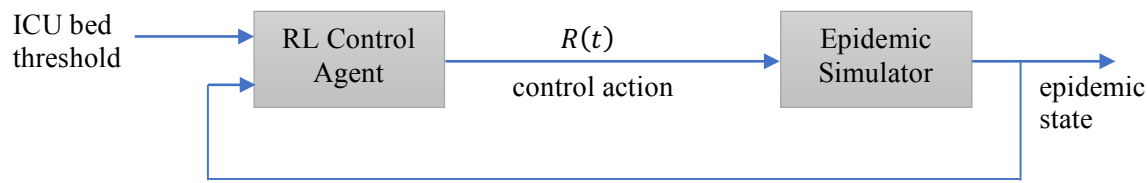

**Figure 1:** Standard RL framework involving RL control agent and environment (epidemic simulator) mapped to a feedback control architecture.

## 1.3. Document Outline

The rest of this document describes the dynamic epidemic model in section 2, the RL control agent and how it was integrated with the epidemic model in section 3, and the experiments performed for cyclic lockdown policy optimization in section 4. Finally, conclusions are summarized in section 5.

## 2. Epidemic Model

To simulate the dynamics of the COVID-19 pandemic we use an extended SEIR (Susceptible, Exposed, Infected, Recovered) model. The SEIR model is a compartmental epidemiological model [15], where the population is divided into groups according to their status with respect to the disease progression. We use an extended SEIR model leveraging the model described in [16]. The compartments are Susceptible (S), Exposed (E), Infected (I), Removed (RM), Mild (M), Severe (SV), Hospitalized (H), Dead (D), and Recovered (RC) as illustrated in Figure 1. The extended model assumes that infected persons self-isolate at the rate of $\gamma$ and move to the Removed (RM) compartment. Removed infected persons can develop either a mild case (with probability $p_m$) where they self-isolate and recover without hospitalization, or a severe case (with probability $p_h = 1 - p_m$) where hospitalization is required and a patient will either recover (with probability $1 - p_{dh}$) or die (with probability $p_{dh}$).

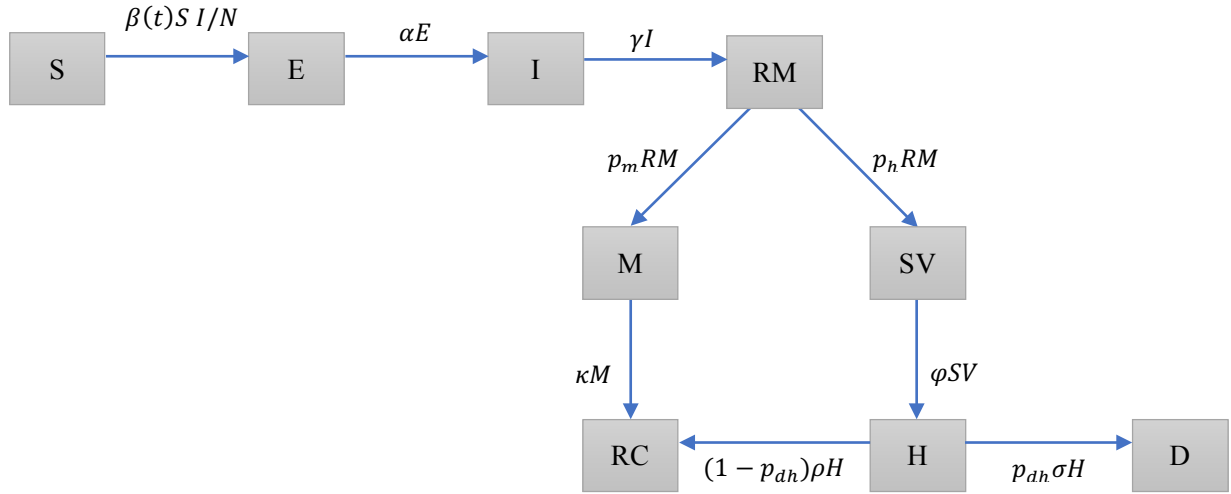

**Figure 2:** Extended SEIR model diagram.

For each of the population groups, its rate of change is modeled as a differential equation and the combined model is a system of differential equations (listed below). In addition to the above probabilities, the differential equations have multiple change rate parameters, whose values are summarized in Table 1. The following is the set of differential equations that define the extended SEIR model:

$$\frac{dS(t)}{dt} = \frac{-\beta(t)S(t)I(t)}{N}$$

$$\frac{dE(t)}{dt} = \frac{\beta(t)S(t)I(t)}{N} - \alpha E(t)$$
$$\frac{dI(t)}{dt} = \alpha E(t) - \gamma I(t)$$
$$\frac{dRM(t)}{dt} = \gamma I(t) - p_m RM(t) - p_h RM(t)$$
$$\frac{dM(t)}{dt} = p_m RM(t) - \kappa M(t)$$
$$\frac{dSV(t)}{dt} = p_h RM(t) - \varphi SV(t)$$
$$\frac{dH(t)}{dt} = \varphi SV(t) - (1 - p_{dh})\rho H(t) - p_{dh}\sigma H(t)$$
$$\frac{dRC(t)}{dt} = \kappa M(t) + (1 - p_{dh})\rho H(t)$$
$$\frac{dD(t)}{dt} = p_{dh}\sigma H(t)$$

**Table 1: extended SEIR model parameter values**

| Parameter | Value | Description | Source |
|---|---|---|---|
| $\beta(t)$ | $\beta(t) = \gamma R(t)$ | Time-dependent transmission rate | |
| $\alpha$ | 1/4 | 1/interval in days for incubation | [3] page 2 |
| $\gamma$ | 1/2 | 1/interval in days from symptoms to removal (isolation) | [16] table 1 |
| $\kappa$ | 1/12 | 1/interval in days for recovery from mild infection | [16] fig. 1 |
| $\varphi$ | 1/5 | 1/interval in days from isolation to hospitalization | [16] fig. 1 |
| $\rho$ | 1/14 | 1/interval in days for recovery from severe infection in hospital | [16] fig. 1 |
| $\sigma$ | 1/14 | 1/interval in days from hospitalization to death | [17] sect. 3.4 |
| $p_h$ | 0.22 | Probability of hospitalization after infected | [1] |
| $p_{dh}$ | 0.17 | Probability of death in hospital after admitted | [2] |

This system of differential equations can be solved through the Euler approximate numerical method as explained in [18]. The Euler method requires converting the set of continuous-time differential equations into a set of discrete-time difference equations that generate estimates of the population values in each of the compartments at each successive time interval: *S(t), E(t), I(t), RM(t), M(t), SV(t), H(t), RC(t), D(t)*. A key variable in this model, as will be described below, is the current number of ICU beds in use, *ICU(t)*. Its value is derived as equal to 30% of the current number of hospitalized persons, *H(t)*.

Our model is dynamic because one of its parameters, the transmission rate, $\beta(t)$, which is the average number of persons an infectious individual transmits the disease to each day, is time-dependent. The transmission rate can be modified through interventions that change the level of interaction between people. The more stringent the social distancing NPIs, the lower the corresponding transmission rate. $\beta(t)$ can also be reduced through interventions that reduce the infectiousness of asymptomatic infected individuals, such as requiring the use of face masks.

The transmission rate is related to $R(t)$, the effective reproduction number, which is the most widely known epidemic metric and represents the severity of an epidemic at any given time. $R(t)$ is defined as the average number of persons one infected individual infects. Mathematically, it is the ratio of the transmission rate over the recovery rate: $\frac{\beta(t)}{\gamma}$. If the transmission rate is larger than the recovery rate, $R(t) > 1$ and it indicates the epidemic is expanding; if the transmission rate is lower than the recovery rate, $R(t) < 1$ and it indicates the epidemic is in decline. We assume the recovery rate is constant and is influenced only by biological and medical factors and not by NPIs. Hence, instead of using $\beta(t)$ as the time-dependent input to the model, we will use $R(t)$, from which $\beta(t)$ is immediately derived as $R(t) \times \gamma$.

---

[1] Based on June 16, 2020 data for New York state: cumulative hospitalized: 90,000, cumulative infected: 405,000. Sources: https://www.worldometers.info/coronavirus/usa/new-york/ and https://www.statista.com/statistics/1109685/new-york-state-covid-cumulative-hospitalizations-us/

[2] Based on https://www.cdc.gov/coronavirus/2019-ncov/hcp/clinical-guidance-management-patients.html: "a range of 26% to 32% of patients were admitted to the ICU" (mean 29%) and "mortality among patients admitted to the ICU ranges from 39% to 72%" (mean 55.5%).

As discussed above, NPIs determine the value of $\beta(t)$ and of $R(t)$. In the Euler method-based solution, the only input to the set of equations on each time interval is an $R(t)$ value (*action*) produced by the RL agent.

## 3. Reinforcement Learning Agent

Reinforcement learning involves a *control agent* that interacts through trial and error with a target system that needs to be controlled and produces as output an optimal *policy function*. The policy function is optimal with respect to goals that are specified as a *reward function*, which is provided as an input to the control agent.

In our case, the target system is the epidemic and obviously it is not safe or realistic to do trial and error interactions with it. However, we use the epidemic model described above as an approximate version of the real system. The policy function produced by the control agent takes as input the state of the system and outputs the best *action* to take at every time step in the progression of the epidemic.

In an online RL setting such this one, the control agent starts with an initial random policy function and interacts with the target system on every time step (daily in our scenario) by observing the current state and applying the policy function to generate an output action. In this scenario, an action is one of two possible $R(t)$ values, corresponding either to non-lockdown (open) conditions or to lockdown (closed) conditions. The model runs for one time step with the new $R(t)$ input, which changes the state of the model. The model's state is the set of compartment population sizes, how many susceptible, infected, hospitalized in ICU, and so on.

The control agent uses a subset of the complete state, referred to as the observed state. With a new observed state, the control agent uses the reward function to calculate the *reward value* triggered by the latest action. On every interaction step, the control agent collects a data tuple comprising: current state, action, next state, and reward value. The tuple is fed to the agent's RL algorithm which incrementally adjusts the policy function so that it maximizes the total sum of reward values in an episode. An episode is a run of the model and the control agent over a sufficient number of steps (days) to observe the evolution of the epidemic.

RL systems need specification of three essential information elements: the *observed state space*, *action space*, and the *reward function*. The observed state space is a collection of metrics that can be observed in the target system and that summarize its current situation from the perspective of the control agent. The observed state space comprises only one variable, the current number of infected persons, *I(t)*. The reason is that for the type of goals that need to be optimized (e.g. operate closest to maximum ICU bed use threshold), this variable can be measured and is sufficient because other variables such as number of hospitalized patients and number of ICU beds in use are dependent on it.

The *action space* is the range of action values that can be produced by the policy function. Recall the action values correspond to $R(t)$ values, which reflect the combined strength of NPIs currently in place. To simplify our analysis, we chose to have only two action values in the action space, a high $R(t)$ value, referred to as $R(t)_{open}$ and a low $R(t)$ value, referred to as $R(t)_{closed}$. The $R(t)_{open}$ value corresponds to non-lockdown NPI combinations (e.g. mandatory use of face masks) and the $R(t)_{closed}$ value corresponds to a lockdown NPI combination (stay-at-home orders, school closures, business closures). Since this a discrete action space, we use an RL algorithm that supports only discrete actions.

## 3.1. Reinforcement Learning Algorithm

The RL algorithm used is Double Deep Q-Network (DDQN) [20]. DDQN is an enhanced version of Q-learning [1] which is a value-based method that iteratively computes a value function, *Q(s, a)*, or *Q-function* whose output is equal to the total value of an action *a* from the current state *s* until the final state of the target system that needs to be controlled. In Q-learning the *Q-function* is defined as an iterative form of the Bellman Equation [1], whose current state, action, next state, and reward inputs are updated on every interaction between the Q-learning agent and the target system. In its most basic form the *Q-function's* mapping of inputs to outputs is represented as a table. However, a table doesn't generalize well to all possible state-action combinations that weren't visited during interaction. To improve generalization, Q-learning replaces the table method with regression as a function approximator. Deep Q-learning refers to methods where the function approximator is a neural network. DQN [19] and DDQN [20] algorithms have significantly improved Q-learning performance through the use of methods including experience replay, mini-batch sampling, and dual networks to contain Q-value overestimation.

## 3.2. Reward Function

The *reward function* defines the goals that are optimized through the RL algorithm. We chose to optimize use of critical resources such as hospital beds, and in particular ICU beds, and to do it with the maximum possible economic activity without exceeding the ICU bed threshold. This is equivalent to maintaining the number of ICU beds in use, *ICU(t)*, as close as possible to the ICU bed threshold, which corresponds to a tracking problem in optimal control. A basic reward function (or cost function when minimizing) for a tracking problem is to use the ICU error value defined as the difference between number of ICU beds in use and the ICU bed threshold to produce a shaped reward, meaning that the higher the error the lower the reward (higher penalty) as expressed in rule (1). The function also uses an error margin whereby a penalty is only applied if the error is lower than the margin.

$$ICU_{error}(t) = ICU_{actual}(t) - ICU_{threshold}$$

$$reward(t) = \begin{cases} 0 & if\ ICU_{error}(t) < margin \\ -\alpha \cdot |ICU_{error}| & if\ ICU_{error}(t) \geq margin \end{cases} \quad (1)$$

This reward function, although simple, produces very unstable results, sometimes rendering a good model and sometimes a very poor model. This instability seems to be caused by a lack of significant differentiation between *Q(s, a)* values with an $R(t)_{open}$ action and *Q(s, a)* values with an $R(t)_{closed}$ action.

To address this issue, we use a reward function that explicitly assigns a portion of its value based on the control input, $R(t)$. This done by dividing the optimization goal into two sub-goals, each with its own reward function and producing a total reward function as a linear combination of the sub-goal rewards functions. The first sub-goal is to operate at the maximum possible level of economic activity, which translates to using $R(t)_{open}$ as much as possible. This is expressed in equation (2) which assigns a reward value of 0 if $R(t)_{open}$ is used, or a penalty (negative value) of $-\alpha_1$, if $R(t)_{closed}$ is used.

The second goal is to avoid overshooting the established ICU bed threshold. This is implemented in equation (3) by assigning a penalty (negative value) if the ICU error is higher than a predefined error margin $\alpha_2$. This penalty is proportional to the ICU error, which is multiplied by a constant set to $\frac{\alpha_1}{\alpha_2}$, so that the resulting penalty value is higher (more negative) than $-\alpha_1$ only for ICU error values higher than $\alpha_2$. The total reward is a linear combination of these two values as expressed in equation (4).

$$reward_1(t) = \begin{cases} 0 & if\ R(t) = R(t)_{high} \\ -\alpha_1 & if\ R(t) \neq R(t)_{high} \end{cases} \quad (2)$$

$$reward_2(t) = \begin{cases} 0 & if\ ICU_{error}(t) \leq \alpha_2 \\ -\dfrac{\alpha_1}{\alpha_2} \cdot ICU_{error}(t) & if\ ICU_{error}(t) > \alpha_2 \end{cases} \quad (3)$$

$$reward_{total}(t) = reward_1(t) + \alpha_3 \cdot reward_2(t) \quad (4)$$

Constants $\alpha_1$ and $\alpha_2$ in the reward equations were selected as follows: $\alpha_1 = 0.1$ and $\alpha_2 = 0.05 \cdot ICU_{threshold}$. $\alpha_3$ is a constant that determines how much weight to assign to the ICU error component of the combined reward function. For each model, it was obtained through manual tuning using multiple trial runs. $\alpha_3$ calculation could be automated by treating it as another input action variable, in addition to $R(t)$, with a method such as described in [21]. This was not pursued in the current work and is left as future work.

### 3.3. Integration of the RL Agent and the Epidemic Simulator

Both the RL agent and the extended SEIR epidemic simulator are implemented in Python and they are integrated using the OpeAI Gym API [22]. This is a simple API that defines a collection of methods and information elements that are needed for interaction between an RL agent and an environment (target system). The main method is *step(a),* which is invoked by the RL agent and executed by the target system. Its execution involves performing action *a* in the simulator and transitioning the system from the current state to the next state. The method's return values are the next state, the reward value, a flag indicating whether or not the end state has been reached, and a field containing any other additional information required by the agent.

## 4. Experiments

The experiments performed involved modeling a hypothetical region with a population of 20 million under the Covid-19 pandemic. The epidemic is simulated from its starting date, assumed to be on March 1, 2020. The initial reproduction number, $R_0$, was assumed to be 3.0 and the simulation runs with this fixed $R$ until a lockdown is applied on March 25. It is assumed the lockdown causes a rapid change in $R$ from 3.0 to 0.7. The lockdown lasts sixty days during which $R$ is fixed at 0.7. At the end of the lockdown, $R$ becomes time sensitive, referred to as $R(t)$ and controlled by the RL agent. After the lockdown, in each step $R(t)$ can switch between two values: $R(t)_{open}$ and $R(t)_{closed}$, which is always set to 0.7. It is assumed that changes from $R(t)_{open}$ to $R(t)_{closed}$ and vice-versa are instantaneous. The minimum duration of an $R(t)$ change is one day.

The purpose of these experiments is to find a policy function for deciding at different $R(t)_{open}$ levels when to impose lockdowns and for how long. The goal of the policy function is to stay as close as possible to an operating threshold or reference value. The operating threshold employed is the maximum ICU beds in use as a percentage of the total ICU capacity. Other resource measurements could also be used as thresholds, such as a percentage of the total number of ICU medical teams available or a percentage of the total number contact tracers. The ICU threshold used is 1,400, which corresponds to 70% of a total ICU capacity of 2,000 units.

The goals of the policy produced by the RL agent are defined by the reward function as described in the previous section. The first goal is to operate at $R(t)_{open}$ as long as possible, which is equivalent to minimizing the time spent in lockdowns (economic goal). The second goal is to avoid overpassing the ICU limit (public health goal). After training against the epidemic model, the RL algorithm produces a near-optimal policy according to these goals. The RL-based results are compared to experiments using the baseline on/off feedback control (fixed-rule) method described in the introduction section.

Lockdown cycles were analyzed for $R(t)_{open}$ values of 1.7, 1.5, 1.3, and 1.1, for both RL-based and baseline on/off control. Figures 3 to 6 illustrate the results of these experiments. Each of the figures has two rows of graphs. The first row (a, b, c) corresponds to a simulation of the epidemic with $R(t)$ as the input control signal generated by the RL agent. The second row (d, e, f) corresponds to a simulation run with $R(t)$ as the input control signal generated by the baseline on/off feedback control method. The first graph in each row shows the number of currently infected persons, *I(t)*, which is used as the RL agent's state variable, the second graph shows the number of occupied ICU beds, *ICU(t)*, and the third graph shows the daily value of the control input $R(t)$. A low $R(t)$, $R(t)_{closed}$, indicates lockdown conditions and a high $R(t)$, $R(t)_{open}$, indicates open conditions. All of the simulations were done over a time span of 270 days, except for $R(t)_{open} = 1.1$, which was done for 365 days.

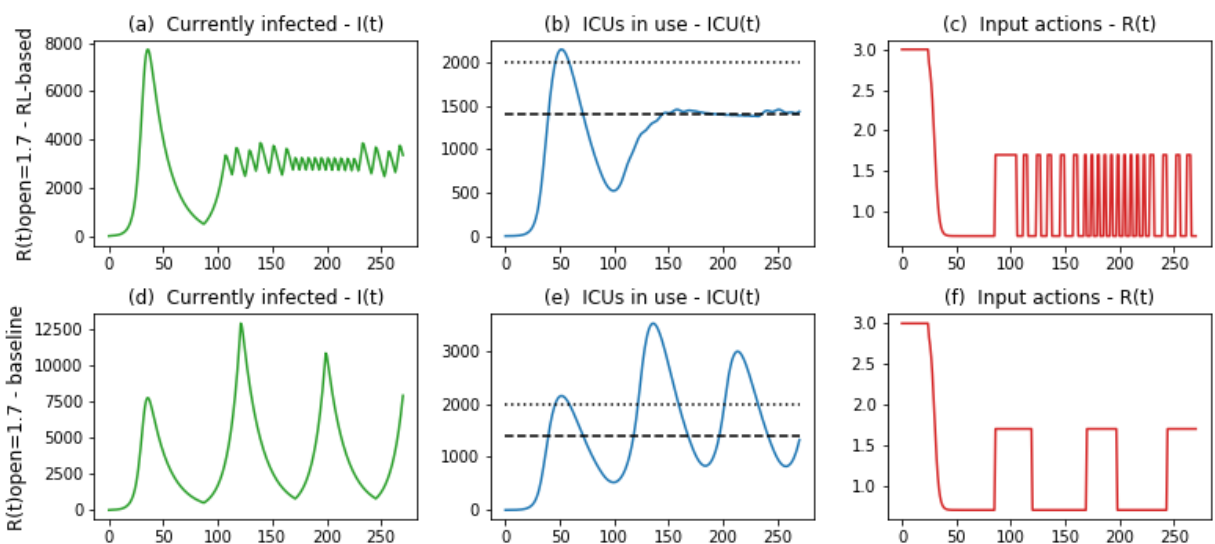

**Figure 3:** RL-based and baseline on/off control with $R(t)_{open} = 1.7$: (a), (b), (c) correspond to RL-based control and (d), (e), (f) to baseline control. *I(t)* is the state variable, *ICU(t)* is the controlled output that needs to track the ICU limit reference value of 1,400. *R(t)* is the control signal produced by the RL agent (c) or the fixed-rule baseline controller (f). Open/closed segment lengths with RL-based control (3c) are 3/6 and with baseline on/off control (3f) are 28/48.

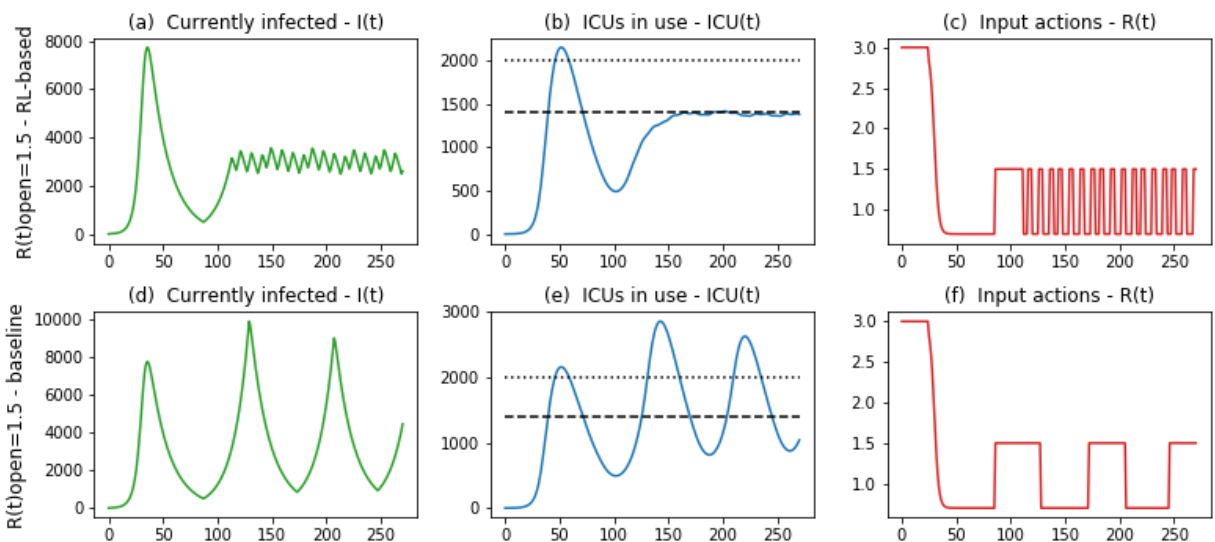

**Figure 4:** RL-based and baseline on/off control with high $R(t)_{open} = 1.5$: Open/closed segment lengths with RL-based control (4c) are 4/6 and with baseline on/off control (4f) are 34/42.

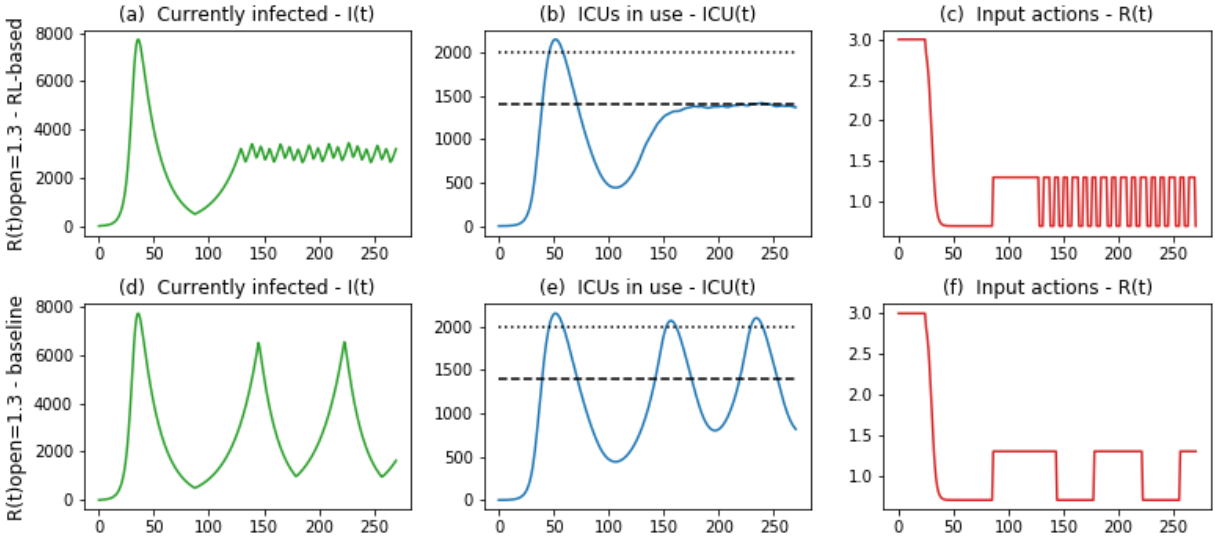

**Figure 5:** RL-based and baseline on/off control with $R(t)_{open} = 1.3$: Open/closed segment lengths with RL-based control (5c) are 5/4 and with baseline on/off control (5f) are 44/34.

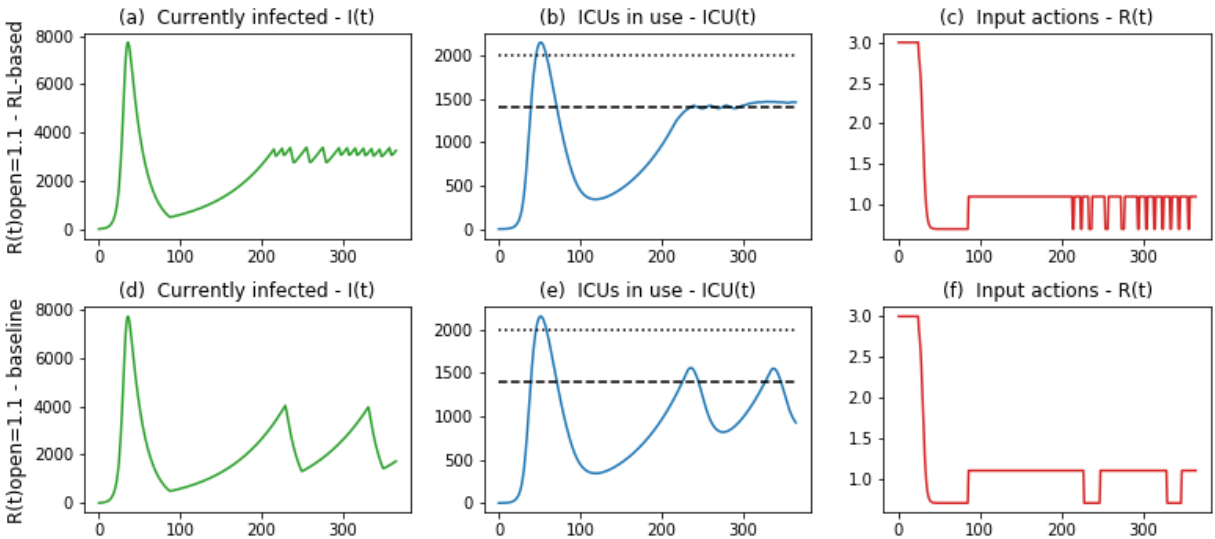

**Figure 6:** RL-based and baseline on/off control with $R(t)_{open} = 1.1$: Open/closed segment lengths with RL-based control (6c) are 11/3 and with baseline on/off control (6f) are 82/19.

As seen in figures 3.b - 6.b, the RL agent succeeds in closely tracking the ICU bed threshold reference value with minimal or no overshoot in every case. In contrast, overshoot values with the baseline on/off feedback control method are very high for $R(t)_{open}$ values above 1.1. With $R(t)_{open} = 1.7$, the maximum overshoot value is 3,520 (Fig. 3e), or 176% the total ICU capacity of 2,000. With $R(t)_{open} = 1.5$ the maximum overshoot value is 2,848 (Fig. 4e), or 142% of the total ICU capacity. With $R(t)_{open} = 1.3$ the maximum overshoot value is 2,100 (Fig. 5e), or 105% of the total ICU capacity. High overshoots with baseline on/off control occur because the epidemic is a dynamic system with inertia in its response to inputs at the system's stages (compartments in the SEIR model). This inertia is caused by response delays to inputs in each of the stages. In the case of ICU beds in use, *ICU(t)*, which is equal to 30% of *H(t)*, requires any change in the input $\beta(t) = \gamma R(t)$ to propagate through the Exposed, Infected, and Severe compartments (see Figure 2) which results in a combined lag of $\frac{1}{\alpha} + \frac{1}{\gamma} + \frac{1}{\varphi}$, which is equal to 11 days based on the model parameters used as listed in Table 1. This means that if a lockdown input is applied, it will only start causing a decrease in *ICU(t)* at least 11 days after the lockdown starts. During this lag interval, *ICU(t)* will continue to grow at an exponential rate dependent on $\beta(t) = \gamma R(t)$. This explains the large increases in offshoots with the baseline on/off control method as $R(t)_{open}$ increases.

Large spikes, as observed in Figures 3e, 4e, and 5e, are undesirable because they imply significantly higher number of deaths given that an increased number of patients in ICU beds results in a larger number of deaths. If spike values are such that they exceed the ICU beds capacity, the situation becomes worse because the fraction of hospitalized patients that die can increase due to lack of medical equipment. The simulation model accounts for this by increasing the probability of death-in-hospital parameter ($p_{dh}$), only during overflow days, by a factor equal to the ratio of ICU beds in use to ICU capacity. For the simulation run with $R(t)_{open} = 1.7$ (Figure 3) when using baseline on/off control, aggregate deaths over a period of 270 days were 19,553, 53% higher than 12,810 deaths when using RL-based control. With $R(t)_{open} = 1.5$ aggregate deaths with baseline on/off control were 15,923, 28% higher, than 12,393 deaths with RL-based control.

The RL agent learns a control policy that minimizes overshoots by applying lockdowns with enough days in advance to either reach very close to the ICU limit or slightly overpass it, but without excess days to avoid an unnecessary undershoot. This results in an *ICU(t)* signal with small oscillations around the ICU limit reference value. The required on/off control signal to produce small *ICU(t)* oscillations needs to have much shorter cycles than cycles in the control signal generated with baseline on/off control or some other blunt control method. The RL agent does learn a policy that produces shorter cycles and its open/closed segment lengths are optimal according to the epidemic model state.

Following, we examine the results for cycles starting at the first cyclic lockdown. For RL-based cycles with $R(t)_{open} = 1.7$ (Figure 3.c) the first cycle starts with a lockdown at day 106 and cycles have average open and closed segment lengths of 3 and 6 days respectively, which amounts 30% of the time open after the first cyclic lockdown. With $R(t)_{open} = 1.5$ (Figure 4.c) the first cyclic lockdown occurs at day 112 and the average open and closed segment lengths are 4 and 6 days, corresponding to 40% of the time open. For $R(t)_{open} = 1.3$ (Figure 5.c) the first cyclic lockdown occurs at day 128 and the average open and closed segment lengths are 5 and 4 days, corresponding to 56% of the time open. For $R(t)_{open} = 1.1$ (Figure 5.c) the first cyclic lockdown occurs at day 214 and the average open and closed segment lengths are 11 and 3 days, corresponding to 79% of the time open.

The key patterns to highlight with RL-based cyclic lockdowns are first, that as the level of virus spread increases, represented by an increase in $R(t)_{open}$, the control policy generates shorter open segment lengths to reduce transmission. The second pattern is that, except for $R(t)_{open} = 1.1$, the total cycle lengths for each $R(t)_{open}$ level are almost the same, between 9 and 10 days and what changes is the open/closed ratio. The total cycle length value depends on the parameters of the epidemic model (compartment delays) and the $R(t)_{open}$ level. We observed two bands in $R(t)_{open}$ levels: $R(t)_{open} \leq 1.1$ and $R(t)_{open} > 1.1$. These two bands result from the exponential growth characteristics of the epidemic and its model. For $R(t)_{open} \leq 1.1$, growth is very slow and considered to be before the elbow of the exponential curve, and for $R(t)_{open} > 1.1$, growth is significantly faster. For $R(t)_{open} \leq 1.1$, the total cycle lengths are slightly higher, such as 13 days in the case $R(t)_{open} = 1.1$.

RL-based cycle lengths ranging from 9 to 13 days are significantly shorter than the cycle lengths with baseline on/off control (see Figures 3f – 6f), which are 76 days long for $R(t)_{open}$ values of 1.7, 1.5, and 1.3, and 101 days for $R(t)_{open} = 1.1$. The ratios of length of open to closed segments are: 28/48, 34/42, 43/33, and 82/19 for $R(t)_{open}$ values of 1.7, 1.5, 1.3, and 1.1 respectively. This indicates that the cycle segment lengths with baseline on/off control are approximately eight times longer than with RL-based control.

As described above, RL-control produces short lockdown cycles because these are the optimal lengths needed to align the controlled variable *ICU(t)* with the tracked reference value (ICU limit). However, in addition to the technical control reasons for using short lockdown cycles there are economic and social reasons to favor short lockdown cycles versus long and stringent lockdowns. Long lockdowns cause severe negative impact on the economy, especially in regions where governments don't have the economic strength to provide adequate help to citizens and businesses affected by the lockdown. Short cyclic lockdowns, as discussed in [6], reduce the negative economic impact because most businesses and informal economic activity could continue functioning by adapting to the short lockdown intervals.

### 4.1. How to apply the RL-based control method

The suggested workflow for applying the RL-based lockdown optimization method is to first measure the current $R(t)$, referred to as $R(t)_{actual}$. This will be the value used for $R(t)_{open}$. The value used for $R(t)_{closed}$ is the lowest $R(t)$ average measured during the initial extended lockdown. Then, the ICU limit value is selected and with these three values as inputs, a new policy model is trained and saved. The RL-agent, then loads the trained policy model and runs it with the simulator for an episode length in days that results in multiple lockdown cycles. Data is recorded as in figures 3c-6c. The policy signal $R(t)$ produced by the RL agent is used to obtain the average lengths of the open and closed segments of a lockdown cycle. These segment lengths are used by a region's government to set up a lockdown plan over multiple cycles.

Once the lockdown plan is in operation, the average $R(t)_{actual}$ over multiple cycles is measured. Then $R(t)_{actual}$ is compared with the calculated average $R(t)_{calc}$ based on the $R(t)_{open}$ and $R(t)_{closed}$ values and on the open and closed segment lengths. For example, for $R(t)_{open}/R(t)_{closed} = \frac{1.7}{0.7}$ and open and closed segment lengths of 3 and 6 days respectively, the calculated average $R(t)_{calc}$ is 1.03. If the measured $R(t)_{actual}$ differs from the $R(t)_{calc}$ by more than a specified margin, then it is an indication that $R(t)_{open}$ needs to be adjusted, assuming $R(t)_{closed}$ remained constant. A new $R(t)_{open}$ can be calculated by solving for it in the $R(t)$ average equation:

$$R(t)_{open} = \frac{cycle_{len} \cdot R(t)_{actual} - closed_{len} \cdot R(t)_{closed}}{open_{len}} \qquad (5)$$

New lockdown cycles are applied with the updated $R(t)_{open}$ and the update process is repeated after a certain number of cycles.

If in addition to lockdowns there are strong parallel efforts to improve other control measures such as testing, contact tracing, isolation, and use of face masks, $R(t)_{open}$ may decrease. This can be determined by comparing $R(t)_{actual}$ measured in the field with the calculated average $R(t)$, $R(t)_{calc}$. If $R(t)_{actual}$ is lower, then a new $R(t)_{open}$ can be calculated with equation (5), a new policy model trained, and new open/closed segment lengths are obtained. When $R(t)_{open}$ decreases, the open/close segment times ratio increases. We refer to this pattern as *downward staircase,* because repeating it enables a stepwise reduction in the percentage of time closed. When reaching $R(t)_{actual} = 0.8$ or lower, $R(t)_{open}$ obtained with equation (5) is 1.0 or lower, which indicates the epidemic is no longer expanding or is contracting, if the value is lower than 1.0, and therefore there is no longer need to perform lockdowns. In summary, the *downward staircase* approach can be used to evolve from short lockdown segments at high $R(t)_{open}$ levels to shorter lockdowns at lower $R(t)_{open}$ levels and eventually to $R(t)_{open}$ levels where no lockdowns are required.

### 5. Conclusions

We developed an approach to calculate optimal cyclic lockdown timings using an RL-based on-off controller. The problem was structured as an optimal control system for tracking a reference value, which in this case is an ICU bed limit. Tracking the ICU limit as close as possible achieves two optimization goals specified in the RL reward function: the first one is a public health goal that minimizes overshoots of ICU bed usage above the ICU bed limit, and the second one is a socio-economic goal that minimizes the time spent under lockdowns.

The RL-based optimal on-off controller succeeds in producing control policies that track the ICU bed limit reference with high accuracy as illustrated in figures 3b-6b. Also, the RL-based controller generates short lockdown cycles, between 9 and 14 days with lockdown intervals of at most 6 days. These results contrast with high ICU limit overshoot values and much longer cycles and lockdown intervals when using basic heuristic feedback methods.

The cyclic lockdown approach described here is intended to only be considered as a temporary alternative to extended lockdowns when a region is facing imminent danger of overpassing resource capacity limits, such as ICU beds, and when imposing an extended lockdown would cause severe social and economic consequences due to the region's lack of necessary economic resources to support its affected population. Under these circumstances, temporary cyclic lockdowns could be used as a bridge method to buy time for improving other processes needed to control the epidemic. When used in conjunction with other methods to reduce the virus spread, cyclic lockdowns can be applied in a downward staircase approach that helps to reduce $R(t)_{open}$ to safe levels close to or lower than 1.0.

Finally, we highlight two areas of improvement and future work. The first one is automating the selection of weight coefficients when the reward function is a linear combination of multiple sub-reward functions, as is the case in the reward function described in section 3.2 and equation (5). This entails exploring methods that treat each of the weights as an additional action variable. The other area is finding optimal controllers for the cyclic lockdown problem using conventional optimal control methods such as model predictive control (MPC) and linear quadratic regulator (LQR) and comparing them with the RL-based solution.